%% file: main.tex
\documentclass[lettersize,journal]{IEEEtran}

\usepackage{amsmath,amsfonts}
\usepackage{algorithmic}
\usepackage{array}
\usepackage[caption=false,font=normalsize,labelfont=sf,textfont=sf]{subfig}
\usepackage{textcomp}
\usepackage{stfloats}
\usepackage{url}
\usepackage{verbatim}
\usepackage{multirow}
\usepackage{graphicx}
\usepackage{booktabs}
\usepackage{tabularx}
\def\BibTeX{{\rm B\kern-.05em{\sc i\kern-.025em b}\kern-.08em
    T\kern-.1667em\lower.7ex\hbox{E}\kern-.125emX}}
\usepackage{balance}
\begin{document}
\title{From ChatGPT to DeepSeek AI: A Comprehensive Analysis of Evolution, Deviation, and Future Implications in AI-Language Models}
 \author{Simrandeep Singh$^1$, Shreya Bansal$^2$,  Abdulmotaleb El Saddik$^3$, Mukesh Saini$^2$\\ $^1$Chandigarh University\\ $^2$Indian Institute of Technology Ropar\\$^3$University of Ottawa

}


\maketitle

\begin{abstract}

The rapid advancement of artificial intelligence (AI) has reshaped the field of natural language processing (NLP), with models like OpenAI’s ChatGPT and DeepSeek AI. Although ChatGPT established a strong foundation for conversational AI, DeepSeek AI introduces significant improvements in architecture, performance, and ethical considerations. This paper presents a detailed analysis of the evolution from ChatGPT to DeepSeek AI, highlighting their technical differences, practical applications, and broader implications for AI development. To assess their capabilities, we conducted a case study using a predefined set of multiple choice questions in various domains, evaluating the strengths and limitations of each model. By examining these aspects, we provide valuable insight into the future trajectory of AI, its potential to transform industries, and key research directions for improving AI-driven language models.

\end{abstract}

\maketitle
\input{1_Introduction}
\input{2_Background}

\input{3_Advancements}
\input{4_Comparison}

\input{5_Implications}

\input{6_Conclusion}
\end{document}

%% file: 1_Introduction.tex
\section{Introduction}
In today’s era, artificial intelligence (AI) is the most significant development in technology; everyone is talking about AI. Its applications are spanning in every field, such as healthcare\cite{rajpurkar2022ai}\cite{johnson2021precision}, robotics\cite{soori2023artificial}\cite{he2021challenges}, finance\cite{cao2022ai}\cite{bahrammirzaee2010comparative}, engineering\cite{naser2023error}\cite{yuksel2023review}, cybersecurity\cite{kaur2023artificial}\cite{khanh2021role},  agriculture\cite{subeesh2021automation}\cite{jha2019comprehensive}, retail\cite{guha2021artificial}, chatbots(Siri, Alexa)\cite{labadze2023role}\cite{lund2023chatgpt}, manufacturing\cite{sahu2021artificial} \cite{li2017applications}, entertainment\cite{hallur2021entertainment}\cite{li2024smart}, business \& marketing\cite{enholm2022artificial}\cite{huang2021strategic}, media\cite{de2021artificial}\cite{ioscote2024artificial}, transportation \cite{wang2023transportation}\cite{abduljabbar2019applications}, and many more. 

AI is helping and facilitating human beings by opening doors for more advanced solutions for the challenges faced by society and pushing the boundaries of conventional methodology to redefine possibilities. AI is a tool derived by computer science engineers to tackle cognitive challenges traditionally associated with human intelligence. It provides solutions for problem-solving, learning, recognizing patterns, summarization, sentiment analysis, chatbots, machine translation, etc. The major agenda of AI is to make the daily life routine of individuals really enjoyable, easy, efficient, convenient, and automated. AI is achieved through machine learning by adopting human-like intelligence and mimicking human behavior, training itself using advanced technologies. It is an essential tool in both practical and entertaining contexts due to its capacity to help humans with a variety of tasks.

One of the sought-after fields in AI is Natural Language Processing (NLP) \cite{otter2020survey}\cite{lane2025natural}, which has become a widely discussed topic after the invention of ChatGPT and similar other tools. However, NLP has several older tools such as ELIZA (1966) \cite{weizenbaum1966eliza}, SHRDLU (1968-1970) \cite{winograd1971procedures}, PARRY (1972) \cite{guzeldere1995dialogues}, LISP-Based NLP Systems (1980s) \cite{foderaro1991lisp}, WordNet (1985-Present) \cite{fellbaum1998wordnet}, Hidden Markov Models (HMM)\cite{rabiner1986introduction}, Latent Semantic Analysis (LSA) (1990s) \cite{landauer1998introduction}, Stanford NLP (2000s-Present), which have paved the way for modern deep learning-based models. 
Human language is a complex phenomenon, having thousands of languages with millions of words and multiple meanings. NLP has emerged as a multidisciplinary field combining AI with linguistics and allows for more significant and realistic communications. NLP can understand, communicate, and interpret language while also facilitating interaction between computers and human language by being trained using machine learning, deep learning, or computational linguistics. NLP includes many steps; after dividing long sentences into individual tokens in tokenization, the position and context of each token are analyzed in tagging. Lemmatization and stemming assist in eliminating affixes and determining the root form of a complete word, which ensures its meaning does not lose contextual flavor. The last phase of processing is chunking, which combines disparate linguistic components into more coherent, structured, and meaningful units \cite{jurafsky2000speech}. 

The introduction of transformer models\cite{han2021transformer}\cite{khan2022transformers} has revolutionized the field of NLP\cite{wolf2020transformers}\cite{gillioz2020overview}. These models, such as GPT\cite{achiam2023gpt}\cite{floridi2020gpt}\cite{liu2024gpt}, have significantly advanced the capabilities of NLP systems, making them more efficient and effective.
Now, machines are becoming more friendly with humans, and models are capable of generating text with human feel and expression. The core of the transformer model is the attention mechanism \cite{vaswani2017attention}\cite{chaudhari2021attentive}, which dynamically gives more attention to key points in the input sequence, making the model capable of tackling sequence-to-sequence tasks, question answering, sentiment analysis, and language modeling with more efficiency. Thus, they can generate new text, understand new patterns and relationships among words, and finally enhance the system's understanding capability. 

One of the major players driving this revolution is OpenAI, established in 2015 as an American artificial intelligence (AI) research lab founded by a group of engineers, researchers, and businesspeople. It has two subsidiary companies—OpenAI Inc. and OpenAI Global LLC—serving non-profit and commercial purposes. The organization has received significant support from well-known individuals and companies, including Microsoft Corporation, Elon Musk, Sam Altman, Ilya Sutskever, and Greg Brockman, who are also co-founders and key investors. The vision behind OpenAI is to develop artificial general intelligence (AGI) \cite{fei2022towards} that surpasses human capabilities, intending to benefit all of humanity. Several machine learning tools, such as DALL-E \cite{ramesh2022hierarchical} and ChatGPT \cite{achiam2023gpt}, have emerged as OpenAI products and are available for public use. ChatGPT, in particular, gained immense popularity, attracting over a million users within just one week of its launch.
OpenAI launched ChatGPT on 30th November 2022, which is based on the GPT-3.5 \cite{brown2020language} and GPT-4\cite{achiam2023gpt} architectures. It has become a widely used innovative tool because of its coherence and versatile applications. It is an advanced chatbot capable of handling a variety of applications such as answering questions, writing code, creating content, providing customer support, assisting with education, drafting emails and meeting minutes, generating ideas, writing project reports, offering healthcare assistance, correcting grammar, conducting research analysis, translating languages, and much more. Its streamlined architecture helps to interpret user input efficiently and provide a response, which mimics real human language. However, ChatGPT possesses many shortcomings, such as high computation cost, a less focused approach, and a higher price point. To resolve these shortcomings, Liang Wenfeng proposed a fresh perspective to NLP models, i.e., Deepseek AI \cite{liu2024deepseek}. 

\begin{table*}[t]
    \centering
     \caption{Comparison of GPT Model Evolution from GPT-1 to GPT-4}
    \renewcommand{\arraystretch}{1.3}
    \begin{tabularx}{\textwidth}{p{4cm}X X X X X}
        \toprule
        \textbf{Feature} & \textbf{GPT-1 (2018)} & \textbf{GPT-2 (2019)} & \textbf{GPT-3 (2020)} & \textbf{GPT-3.5 (2022)} & \textbf{GPT-4 (2023)} \\
        \midrule
        Parameters & 117 million & 1.5 billion & 175 billion & 200-300 billion & Estimated 1T+ (not officially disclosed) \\
        Training Data & BooksCorpus $\sim$7K books & WebText (8M webpages) & 570GB of text from books, articles, and the internet & Improved over GPT-3 with better filtering & Vast collection of data scraped from the internet, including books, websites, scientific papers, etc. \\
        Context Length & $\sim$512 tokens & $\sim$1024 tokens & $\sim$2048 tokens & $\sim$4096 tokens & $\sim$32K \& $\sim$128K tokens in GPT-4 Turbo \\
        Transformer Layers & 12 & 48 & 96 & Similar to GPT-3 & Estimated 100+ \\
        Modality & Text only & Text only & Text only & Text only & Text + Images (Multimodal) \\
        Multilingual Support & Limited English & Basic multilingual understanding & Supports multiple languages but mainly trained in English & Better non-English understanding & Strong multilingual capabilities (supports 25+ languages well) \\
        Few-shot Learning & No & Partial & Yes & Improved & Advanced Few-shot \& Zero-shot learning \\
        Logical Reasoning & Weak & Moderate & Better, but inconsistent & Improved, but still flawed & Strongest yet, closer to human-level reasoning \\
        Performance on Benchmarks & Low & Moderate & High & Higher & Best so far (passes simulated bar exam, high SAT/GRE scores, etc.) \\
        Creativity & Low & Moderate & High & Higher & Best for creative writing, storytelling, and code generation \\
        Factual Accuracy & Poor & Moderate & Often hallucinates & Fewer hallucinations & Most reliable, fewer hallucinations \\
        Computation Cost & Low & High & Very high & Optimized over GPT-3 & Very high, but optimized efficiency \\
        Internet Access & No & No & No & No & No direct access, but trained on a larger dataset \\
        Fine-tuning Capability & Limited & Somewhat customizable & Available for enterprises & More customizable & Advanced fine-tuning support \\
        Code Generation & Very basic & Improved & Strong (GPT-3 Codex used in GitHub Copilot) & Even better & Best for programming, used in AI coding tools \\
        Bias \& Ethical Issues & High & Still significant & Moderate, but problematic & Improved with better moderation & Best moderation \& bias reduction \\
        Accessibility & Research only & Open to public (some restrictions) & Commercial API (GPT-3.5 Turbo made it cheaper) & API \& ChatGPT integration & ChatGPT-4 available via API and subscription \\
        Cost & Low & High (due to more parameters) & Very high (expensive inference) & More cost-effective than GPT-3 & GPT-4 Turbo made it cheaper and faster \\
        \bottomrule
    \end{tabularx}  
   
    \label{tab:gpt_comparison}
\end{table*}

%% file: 2_Background.tex
\section{Background}
This section reviews the evolution of ChatGPT, highlighting its development and capabilities across different versions. It also introduces DeepSeek AI, a new approach that aims to address some of the limitations of current models like ChatGPT, offering a more efficient and task-focused paradigm for NLP.

\subsection{ ChatGPT: A Pioneering Model}
ChatGPT is a publicly available AI tool developed by OpenAI, marking a significant advancement in natural language processing (NLP) and conversational AI. The basic building block of ChatGPT is a large language model (LLM) \cite{chang2024survey}\cite{liang2024survey} architecture, which includes embedding, encoder-decoder layers \cite{badrinarayanan2017segnet}, positional encoding, self-attention mechanisms, feed-forward networks, add \& normalization layers, and multi-head attention. ChatGPT is a highly sophisticated chatbot implemented through a deep neural network architecture using a transformer framework to generate coherent and contextually relevant text. It belongs to a group of widely used transformer-based models including Bidirectional Encoder Representations from Transformers (BERT), and Generative Pre-trained Transformers (GPT). ChatGPT is a language model that comprehends human-like text across a wide range of applications such as sentence completion, translation, and conversational interaction. It simulates conversations with human users and generates human-like outputs.  Its conversational abilities are enhanced using fine-tuning conducted by reinforcement learning with human feedback (RLHF) \cite{griffith2013policy}. The capabilities of this model are enhanced using extensive pre-training conducted using diverse datasets sourced from various books, articles, websites, and other textual content.  utilizing high-end GPUs.

\subsection{Evolution of ChatGPT}
Large language models (LLMs) act as a foundation stone for the rapid evolution of natural language processing (NLP). After the significant advancement of LLM, ChatGPT has undergone multiple iterations in a short period. Within 2 years only, ChatGPT has quickly improved its capabilities and performance parameters. The groundbreaking transformer models laid the foundation for more advanced large language models. These transformer models have demonstrated significant improvements over traditional Recurrent Neural Networks (RNN) and Long Short-Term Memory (LSTM) \cite{sherstinsky2020fundamentals} models. The encoder-decoder architecture of transformers has proven a seismic shift in the deep learning horizon. As illustrated in Fig. 2, the transformer model shows parallel processing abilities and is trained to understand and generate human-like text. RNNs and LSTMs face challenges such as vanishing gradients \cite{hochreiter1998vanishing} when dealing with long dependencies. However, transformers excel in parallel processing due to their reliance on the `attention mechanism', which enables them to capture relationships across long data sequences. 
The first GPT-1 model \cite{radford2018improving}, introduced in mid-2018, utilized auto-regressive language modeling as an unsupervised pre-training approach. This approach set the foundation for pre-training on large text corpora followed by fine-tuning, becoming a standard methodology for various NLP tasks. During the pre-training phase, the GPT model uses a traditional language modeling objective, as illustrated in Eq. \ref{eq:GPT1}:
\begin{equation}
    L_1 (u) = \sum_{i} \log P(u_i \mid u_{i-k}, \dots, u_{i-1})
    \label{eq:GPT1}
\end{equation}
where \( u_i \) is the current token, \( u_{i-1} \), \( u_{i-2} \) \(\dots\) \( u_{i-k} \) are the previous  \( k \) context tokens, and \( P \) is the probability function modeled using a decoder-only transformer. After pre-training, the model is fine-tuned for specific tasks through supervised learning, where it is trained on relevant datasets with input transformations. During inference, GPT-1 generates new sequences, utilizing the \(117\) million parameters it was trained on. The training process involved processing around \(7,000\) unpublished books. 

OpenAI released GPT-2 in \(2019\) \cite{radford2019language}, featuring a \(1.5\) billion-parameter transformer. The model includes parameters such as a vocabulary size of over \(50,000\), \(12\) attention heads, \(12\) layers, and a batch size of 512. It was trained on 8 million web texts or web pages, without the need for supervised fine-tuning. A notable feature of GPT is its ability to perform zero-shot learning, enabling it to handle tasks it has not been explicitly trained for. This capability is achieved by leveraging patterns and knowledge acquired during training to generalize across unseen tasks. For instance, the model can classify sentiment or generate creative content without requiring specific task-related examples in its training data. Language modeling used in GPT 2 is given by Eq. \ref{eq:GPT2}, which represents the probabilistic framework for the probability of a sequence \( u_i \) given its preceding states  \( u_{i-1} \), modeled as a product of conditional probabilities.
\begin{equation}
    p(x) = \prod_{i=1}^{n} P(u_i \mid u_1, \dots, u_{n-1})
    \label{eq:GPT2}
\end{equation}

The architecture of GPT-3 \cite{brown2020language} doesn’t have much variation as compared to GPT-2, the key change carried in GPT-3 is the use of alternating dense and locally banded sparse attention patterns within the transformer framework. This extensive dataset, comprising about 410 billion tokens, allowed the autoregressive language model (GPT-3) to develop a broad understanding of language patterns and contextual relationships and was introduced in 2020. Training was carried out on the huge data, comprising of approximately 570GB of text after filtering, sourced from Common Crawl (60\% of the training mix), WebText2 (19 billion, 22\% training weight), Books1 (19 billion, 8\% training weight), Books2 (55 billion, 8\% training weight), and Wikipedia (3 billion, 2\% training weight). GPT-3 is utilized without gradient updating or fine-tuning, based on tasks and a small set of demonstrations specified only in terms of text interaction with the model. GPT-3 is a few-shot and multitask model trained on 8 models of different sizes, having trainable parameters ranging between 125M to 175B. GPT-3 is 10x more advanced than previous versions and has wide applications such as language translation, content creation, text classification, sentiment extraction, creative writing, writing assistance, research and analysis, generating code, business guidelines, and more.
GPT-3.5 is just a fine-tuned and iterated version of GPT-3, introduced in the year 2022. It is capable of generating more realistic, relative, and coherent text as compared to previous versions. The parameters of GPT-3 and 3.5 	have increased significantly, representing a substantial improvement over earlier versions. OpenAI launched ChatGPT in 2023 and the foundation model is GPT-3.5. GPT-3.5 is capable of generating human-like text known as humanized AI, showing deeper knowledge of the semantics and context of text, and hence enabling it to perform better for technical and report writing. 
\begin{figure}
    \centering
    \includegraphics[width=1\linewidth]{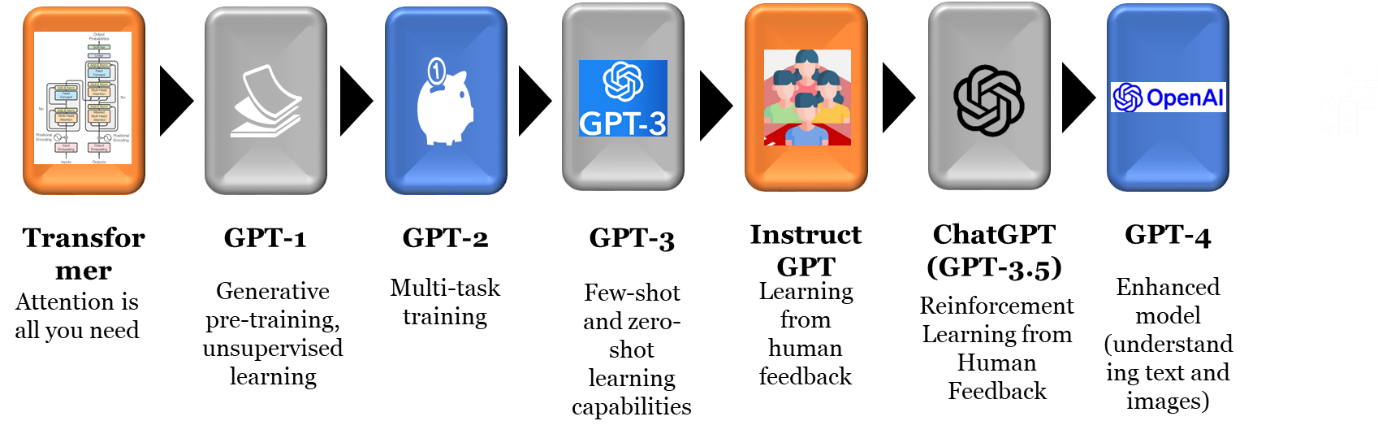}
    \caption{Evolution of GPT: From Version 1 to 4}
    \label{fig:enter-label}
\end{figure}

GPT-4 \cite{achiam2023gpt} entered the public domain on March 14, 2023, with improved reasoning ability. It shows multimodal behavior, i.e. compatible with both text and images as inputs. This behavior enables GPT-4 to understand visuals, spoken words, and text information, which has enhanced its ability to respond to complex and comprehend long-term contexts. This significant improvement has enabled GPT-4 to store longer versions of data, preserve details throughout the conversation, and provide more ethical and fair outputs. All these developments are summarized in Table \ref{tab:gpt_comparison}, which highlights the key differences and advancements across each GPT iteration.

\subsection{DeepSeek AI: A Paradigm Shift
}

DeepSeek AI, developed by DeepSeek, builds on the foundation of ChatGPT but introduces significant innovations. It comes with the motive to enhance Artificial General Intelligence (AGI) and to make it a reality. It includes advanced fine-tuning techniques, a deeper focus on contextual understanding, Graph Neural Networks (GNNs) \cite{wu2020comprehensive}, Reinforcement Learning, or Memory-Augmented Networks \cite{santoro2016meta}, and a focus on ethical AI practices. DeepSeek AI has been declared to be more domain-specific and aims to overcome the limitations of ChatGPT. DeepSeek is using the model with optimized efficiency, reducing biases, and providing more customized responses. The clear agenda of its development seems to shift toward more responsible and adaptable AI systems.

%% file: 3_Advancements.tex
\section{Key Deviations and Advancements}
The Chinese AI research lab established in 2023 developed the fully open-source DeepSeek R1 model and launched it for the public in 2025. It is getting significant attention worldwide due to its cost-effective training. 
It varies from its counterpart in terms of reasoning and non-reasoning capabilities, such as self-verification, reflection, and long conversations. On the architecture level, it replaces supervised fine-tuning with reinforcement learning (RL), a training pipeline involving two RL stages, and two supervised fine-tuning (SFT) stages.

DeepSeek-R1-Zero represents a novel approach in LLM model, RL directly applied to the base model and bypassing the traditional supervised fine-tuning (SFT) stage. This innovative method enables the model to autonomously explore and develop chain-of-thought (CoT) reasoning strategies for tackling complex problems.
This training approach helps DeepSeek-R1-Zero to achieve significant advancements in AI, such as self-verification, reflection, and the generation of extensive chains of thought. Self-verification helps to assess and validate its own outputs, and reflection presents an introspective analysis of its reasoning process.
It demonstrates the capabilities of RL-based training to foster more sophisticated and self-aware language models.

Additionally, DeepSeek's performance has improved using model distillation \cite{gou2021knowledge}, which enables smaller models to achieve the reasoning abilities of larger models. The total training cost is significantly lower than that of other renowned LLM models like Google and OpenAI, which have spent much more on similar foundation models. The cost per inference is also much lower, making it an attractive option for scalable deployment.

Chinese AI research lab has utilized H800 chips, employing techniques like mixture-of-experts and multi-head latent attention to compensate for lower computational power. This breakthrough allowed the model to perform effectively despite hardware constraints. The following subsections outline its architectural improvements, performance metrics, ethical considerations, and practical applications.
\subsection{ Architectural Improvements}
\begin{enumerate}
\item{Model Size and Efficiency: ChatGPT relies on a massive number of parameters (e.g., 175 billion in GPT-3), which contributes to its high computational costs. DeepSeek AI, on the other hand, employs a more efficient architecture, reducing parameter counts while maintaining or even improving performance. This is achieved through techniques like sparse attention mechanisms and model distillation.}
\item{Fine-Tuning and Adaptability: DeepSeek AI incorporates advanced fine-tuning methods, such as reinforcement learning from human feedback (RLHF) and domain-specific pre-training. This allows the model to adapt more effectively to specialized tasks, such as medical diagnosis or legal document analysis.}
\item Group Relative Policy Optimization (GRPO): To reduce the computational expenses associated with reinforcement learning (RL), DeepSeek employs Group Relative Policy Optimization (GRPO), a method introduced by Shao et al. \cite{shao2024deepseekmath} in 2024. GRPO is an online learning algorithm that offers a more efficient alternative to traditional approaches by eliminating the need for a separate critic model. GRPO aims to maximize the advantage of the generated completions that help a model to learn better by comparing different actions and making small, controlled updates using a group of observations.

 Instead of using a critic, GRPO employs a group-based evaluation strategy. The algorithm generates multiple outputs from the existing policy ($\pi_{\theta_{\text{old}}}$) for each given question or prompt. It then uses these outputs to establish a baseline for performance evaluation.
The optimization process for the policy model ($\pi_{\theta}$) involves maximizing an objective function that compares the relative performance of outputs within each group. This approach allows for a more streamlined and cost-effective training process while maintaining the ability to improve the model's performance.
The group reinforcement learning objective (GRPO) is defined by Eq. \ref{equ:grpo}.

\begin{equation}
    L_{\text{GRPO}}(\theta) = L_{\text{clip}}(\theta) - w_1 D_{\text{KL}}(\pi_{\theta} \| \pi_{\text{orig}})
\label{equ:grpo}
\end{equation}

where:  
\begin{itemize}
    \item $L_{\text{clip}}(\theta)$ is the clipped surrogate loss, similar to PPO.
    \item $D_{\text{KL}}(\pi_{\theta} \| \pi_{\text{orig}})$ is the KL divergence term.
    \item $w_1$ is a weight parameter.
\end{itemize}

The advantage for each response in a group is calculated by Eq. \ref{equ:response}:

\begin{equation}
    A_i = \frac{R_{\phi}(r_i) - \text{mean}(G)}{\text{std}(G)}
\label{equ:response}
\end{equation}

Where:  
\begin{itemize}
    \item $R_{\phi}(r_i)$ is the reward for response $r_i$.
    \item $G$ is the group of responses.
    \item $std$ is the standard deviation.
\end{itemize}

\end{enumerate}

\subsection{Performance Metrics
}
\begin{enumerate}
   \item Contextual Understanding: One of ChatGPT's limitations is its tendency to lose context in long conversations. DeepSeek AI addresses this by implementing memory-augmented architectures, enabling it to maintain coherence over extended interactions.
\item Bias Mitigation: ChatGPT has been criticized for generating biased or inappropriate content due to biases in its training data. DeepSeek AI employs debiasing algorithms and curated datasets to minimize such occurrences, ensuring more equitable and responsible outputs.
\item Multilingual Capabilities: While ChatGPT supports multiple languages, DeepSeek AI enhances this capability by incorporating low-resource languages and improving translation accuracy through cross-lingual transfer learning.

\end{enumerate}

\subsection{Ethical Considerations}

\begin{enumerate}
\item Transparency and Explainability: DeepSeek AI prioritizes transparency by providing users with insights into how responses are generated and mechanisms to improve response quality over time. This includes explainable AI (XAI) \cite{ali2023explainable}\cite{dwivedi2023explainable} techniques such as SHAP (Shapley additive explanations) \cite{lundberg2017unified} and LIME (Local interpretable model-agnostic explanations)\cite{ribeiro2016should} that highlight the reasoning behind specific outputs. It doesn't use a black box mechanism; instead, the decision-making process is made traceable and auditable.
\item User Privacy: DeepSeek AI incorporates cutting-edge privacy-preserving measures, such as differential privacy\cite{abadi2016deep} and federated learning\cite{mcmahan2017communication}, to ensure that user interactions remain confidential. It protects user data and unauthorized surveillance by adding mathematical noise to the signal. Federated or collaborative learning ensures the training of models on the local hardware and transmits weights and biases to a central server for improving the global model. 
\item Ethical Alignment: DeepSeek AI is designed with ethical guidelines such as fairness, accountability, and inclusivity embedded into its training process, reducing the risk of harmful or unethical outputs.
DeepSeek models undergo rigorous bias detection using fairness-aware algorithms to detect harmful, adult, misleading, or offensive content. The model is improved for ethical consideration using reinforcement learning with human feedback (RLHF) to improve ethical decision-making over time.
\end{enumerate}

\subsection{Practical Applications}
\begin{enumerate}
\item Industry-Specific Solutions: DeepSeek AI offers tailored solutions for various industries. For example, in healthcare, it can assist with medical diagnosis and patient communication. At the same time, it can analyze market trends, risk assessment, investment decisions, fraud detection, customer service and generate reports in finance. Retail market, education, and autonomous systems are other industries where DeepSeek is transforming conventional techniques. 
\item Real-Time Adaptability: Unlike ChatGPT, which operates primarily in a static manner, DeepSeek AI can adapt to real-time changes in input, making it suitable for dynamic environments such as live customer support or interactive education. It is best suited for applications where immediate response is required, such as traffic adaptive traffic light signals, detecting fraudulent transactions in finance, personalized tutoring according to the aptitude level of students, report generation for MRI, CT, and other scans,  real-time sentiment analysis, and market trends.
\item Creative Applications: DeepSeek AI's enhanced creativity and coherence make it a valuable tool for content creation, including writing, music composition, and graphic design. It finds vast applications in the areas such as drafting content, brainstorming, or writing code, take notes, making minutes of meetings; discovering complex reasoning patterns, interactive storytelling, simulations for complex systems, idea generation, design optimization and many more.

\end{enumerate}

%% file: 4_Comparison.tex
\section{Comparative Analysis}
DeepSeek is also open-sourced, promoting competition and encouraging further advancements in AI development. This could lead to reduced costs and better models in the future, benefiting companies and users worldwide.
This section provides a detailed comparison of ChatGPT and DeepSeek AI across several dimensions such as model architecture, training data and methodology, reinforcement learning, computational efficiency, context, and ethical, and societal implications along with summarized Tables. \ref{tab:DeepSeek_gpt_comparison} and \ref{tab:chatgpt_vs_deepseek}. Here deeper discussion and insights of training, capabilities, and limitations are presented.

\begin{enumerate}
    \item Model Architecture 
    \begin{itemize}
\item ChatGPT: It is a general-purpose language model based on OpenAI’s GPT-3.5 or GPT-4 architecture, employing a dense Transformer model with a focus on large-scale pre-training and fine-tuning using Reinforcement Learning from Human Feedback (RLHF). It undergoes large-scale pretraining and offers high computational cost with high latency
\item DeepSeek AI: It utilizes an optimized and hybrid Transformer architecture with enhanced attention and active learning mechanisms, improving context retention and reducing token dependencies for better long-form coherence. It is best suited for potentially domain-specific or task-optimized applications. 

\end{itemize}
\begin{table*}[t]
    \centering
    \renewcommand{\arraystretch}{1.3}
    \caption{Comparison of DeepSeek AI and GPT Series (GPT-1 to GPT-4)}
    \begin{tabularx}{\textwidth}{p{4cm}X X}
        \toprule
        \textbf{Feature} & \textbf{DeepSeek AI} & \textbf{GPT Series (GPT-1 to GPT-4)} \\
        \midrule
\textbf{Parameters}            & Likely in the range of tens to hundreds of billions (exact number undisclosed) & Ranges from 117M (GPT-1) to 1T+ (GPT-4).                                                       \\ 
\textbf{Training Data}         & Large-scale, diverse datasets, possibly including multilingual and multimodal data. & Evolved from BooksCorpus (7K books) to a massive, diverse dataset including text and images.   \\ 
\textbf{Context Length}        & Likely competitive with GPT-4 (e.g., 128K tokens or more).                      & Improved from 512 tokens (GPT-1) to 128K tokens (GPT-4 Turbo).                                 \\ 
\textbf{Transformer Layers}    & Likely similar to GPT-4 (100+ layers).                                          & Increased from 12 layers (GPT-1) to 100+ layers (GPT-4).                                       \\ 
\textbf{Modality}              & Likely multimodal (text + images + potentially other modalities).               & GPT-4 introduced multimodal capabilities (text + images).                                      \\ 
\textbf{Multilingual Support}  & Strong multilingual capabilities, possibly supporting 25+ languages.            & Improved from limited English (GPT-1) to strong multilingual support (GPT-4).                 \\ 
\textbf{Few-shot Learning}     & Advanced few-shot and zero-shot learning capabilities.                          & Improved from none (GPT-1) to advanced few-shot and zero-shot learning (GPT-4).               \\
\textbf{Logical Reasoning}     & Strong logical reasoning, potentially competitive with GPT-4.                   & Improved from weak (GPT-1) to human-level reasoning (GPT-4).                                  \\ 
\textbf{Performance on Benchmarks} & Likely competitive with GPT-4 on standard benchmarks.                          & Improved from low (GPT-1) to best-in-class (GPT-4).                                           \\ 
\textbf{Creativity}            & High creativity in text generation, storytelling, and code generation.          & Improved from low (GPT-1) to best-in-class (GPT-4).                                           \\ 
\textbf{Factual Accuracy}      & Improved factual accuracy with fewer hallucinations.                            & Improved from poor (GPT-1) to most reliable (GPT-4).                                          \\ 
\textbf{Computation Cost}      & Likely high but optimized for efficiency.                                       & Increased from low (GPT-1) to very high but optimized (GPT-4).                                \\ 
\textbf{Internet Access}       & No direct access, but trained on up-to-date datasets.                           & No direct access, but GPT-4 trained on a larger, more recent dataset.                         \\ 
\textbf{Fine-tuning Capability} & Advanced fine-tuning support for enterprises.                                   & Improved from limited (GPT-1) to advanced fine-tuning support (GPT-4).                        \\ 
\textbf{Code Generation}       & Strong code generation capabilities, possibly competitive with GPT-4.           & Improved from very basic (GPT-1) to best-in-class (GPT-4).                                    \\ 
\textbf{Bias \& Ethical Issues} & Likely improved moderation and bias reduction.                                  & Improved from high (GPT-1) to best moderation and bias reduction (GPT-4).                     \\ 
\textbf{Accessibility}         & Likely available via API and subscription models.                               & Improved from research-only (GPT-1) to API and subscription models (GPT-4).                   \\ 
\textbf{Cost}                  & Likely competitive with GPT-4 Turbo in terms of cost-effectiveness.             & Improved from low (GPT-1) to cost-effective (GPT-4 Turbo).                                    \\
        \bottomrule
    \end{tabularx}  
    
    \label{tab:DeepSeek_gpt_comparison}
\end{table*}
    \item Training Data and Methodology

    \begin{itemize}
        \item ChatGPT: Trained on a diverse dataset, including internet text, books, and academic papers, with additional fine-tuning through RLHF.
\item DeepSeek AI: Employs a more dynamic dataset integration approach, incorporating real-time updates and domain-specific datasets for improved adaptability in specialized fields.
    \end{itemize}

    \item Reinforcement Learning and Optimization

    \begin{itemize}
        \item ChatGPT: Uses RLHF to refine responses and improve user alignment, focusing on reducing biases and enhancing conversational relevance.
\item DeepSeek AI: Advances RLHF with dynamic reinforcement mechanisms, incorporating adaptive reward modeling and improved human-AI feedback loops for more fine-tuned responses.
    \end{itemize}

    \item Computational Efficiency and Scalability
    \begin{itemize}
        \item ChatGPT: Requires significant computational resources due to its dense architecture and extensive training cycles.
\item DeepSeek AI: Employs model compression techniques such as knowledge distillation and quantization to optimize performance and reduce computational overhead.
    \end{itemize}

    \item Context Window and Memory Retention
\begin{itemize}
        \item ChatGPT: Supports a large but fixed context window, limiting its ability to recall previous interactions beyond a certain token limit.
\item DeepSeek AI: Implements an improved context window management system, allowing better retention of conversational history across longer interactions.
    \end{itemize}

    \item Societal and Ethical Implications
\begin{itemize} 
\item Bias and Fairness: Both models face challenges related to bias in AI-generated content. DeepSeek AI’s emphasis on domain-specific customization offers potential for greater fairness but also introduces risks of overfitting to specific viewpoints.

\item Impact on the Workforce: AI language models are increasingly influencing industries such as content creation, customer support, and programming. While they enhance productivity, they also raise concerns about job displacement and the need for new skill sets.
  \end{itemize}

\item Ethical Considerations and Future Regulation: As AI becomes more pervasive, regulatory frameworks will play a crucial role in mitigating misuse. Transparency in training methodologies and responsible AI deployment remain key areas of discussion.
\end{enumerate}

\begin{table*}[t]
    \centering
    \caption{Comparison between ChatGPT and DeepSeek AI}
    \renewcommand{\arraystretch}{1.3}
    \begin{tabularx}{\textwidth}{p{4cm}X X} 
        \toprule
        \textbf{Dimension} & \textbf{ChatGPT} & \textbf{DeepSeek AI} \\
        \midrule
        Architecture & Transformer-based, large parameter count & Optimized architecture, fewer parameters \\
        Fine-Tuning & General-purpose fine-tuning & Domain-specific fine-tuning \\
        Contextual Understanding & Limited in long conversations & Enhanced with memory-augmented systems \\
        Bias Mitigation & Limited debiasing techniques & Advanced debiasing algorithms \\
        Ethical Alignment & Basic ethical guidelines & Embedded ethical frameworks \\
        Computational Efficiency & High computational costs & Optimized for efficiency \\
        Real-Time Adaptability & Limited & High \\
        \bottomrule
    \end{tabularx} 
    
    \label{tab:chatgpt_vs_deepseek}
\end{table*}

\subsection{Case study}
We have conducted a comprehensive evaluation assessment of ChatGPT and DeepSeek models by asking a predefined set of multiple-choice questions spanning various domains. The results of the comparative case study are represented in Table \ref{case study}. It evaluates the performance capabilities of ChatGPT and DeepSeek across 24 domains using multiple-choice questions. The table reports the number of Total Questions posed to each model, along with the number of Total Correct answers and the corresponding Accuracy (\%) for both ChatGPT and DeepSeek.

Overall, DeepSeek AI outperforms ChatGPT in terms of accuracy across most domains. For example, in the tourism domain, DeepSeek AI correctly answered 85 of 100 questions, resulting in an accuracy of 85\%, while ChatGPT correctly answered 53 of 100 questions, with an accuracy of 53\%. Similarly, in the Physics domain, DeepSeek AI achieved 92\% accuracy, correctly answering 46 out of 50 questions, while ChatGPT answered 43 out of 50 questions, achieving 86\% accuracy.

However, there are domains where ChatGPT performed equally or better than DeepSeek AI. For example, in Psychology and Economics, both models achieved perfect accuracy, answering all questions correctly (100\%). In domains like Mechanical Engineering, Botany, and Commerce, the performance of both models are more comparable. For example, in Mechanical Engineering, both ChatGPT and DeepSeek AI correctly answered 39 of 50 questions, resulting in an accuracy of 78\% for both.

In mathematics, DeepSeek AI performed better, achieving perfect accuracy by answering all 53 questions correctly (100\%), while ChatGPT answered 43 of 53 questions, resulting in an accuracy of 81\%. This shows that DeepSeek AI outperforms ChatGPT in mathematics by achieving higher accuracy. Similarly, in Commerce, DeepSeek AI outperformed ChatGPT, answering 49 out of 50 questions correctly (98\% accuracy) compared to ChatGPT's 42 correct answers (84\% accuracy).

A comprehensive summary of the overall performance of both models is provided at the bottom of the table. Across all 1429 questions tested, DeepSeek AI answered 1245 questions correctly, achieving an overall accuracy of 87\%. In comparison, ChatGPT answered 1140 questions correctly, with a total accuracy of 79\%. This overall performance reinforces the trend that DeepSeek AI generally outperforms ChatGPT in terms of accuracy across a wide range of domains.

In summary, while ChatGPT performs well in many domains, DeepSeek AI consistently delivers higher accuracy in most cases, with notable exceptions like Psychology and Economics where both models perform equally. DeepSeek AI also shows particular strength in domains like Mathematics, where it achieved perfect accuracy, while ChatGPT's accuracy was lower.

\begin{table*}[ht]
    \centering  
    \caption{Case Study: Performance comparison of ChatGPT and DeepSeek multiple choice questions across various domains}
    \label{case study}
    \renewcommand{\arraystretch}{1.3}
    \begin{tabularx}{\textwidth}{p{3.5cm}p{2.5cm}p{2.5cm}p{2.5cm}p{2.5cm}p{2.5cm}} 
        \toprule
        \multicolumn{1}{c}{\multirow{2}{*}{Domain}} & \multicolumn{1}{c}{\multirow{2}{*}{Total Questions}} & \multicolumn{2}{c}{ChatGPT} & \multicolumn{2}{c}{DeepSeek} \\
        \multicolumn{1}{c}{}                        & \multicolumn{1}{c}{}                                 & Total Correct   & Accuracy (\%)  & Total Correct   & Accuracy (\%)   \\
        \midrule
        Tourism                                     & 100                                                  & 53              & 53\%      & 85              & 85\%       \\
        Psychology                                  & 50                                                   & 50              & 100\%     & 50              & 100\%      \\
        Physics                                     & 50                                                   & 43              & 86\%      & 46              & 92\%       \\
        Mechanical                                  & 50                                                   & 39              & 78\%      & 39              & 78\%       \\
        Mathematics                                 & 53                                                   & 43              & 81\%      & 53              & 100\%      \\
        English                                     & 101                                                  & 64              & 63\%      & 78              & 77\%       \\
        CSE                                         & 50                                                   & 49              & 98\%      & 48              & 96\%       \\
        ECE                                         & 55                                                   & 48              & 87\%      & 50              & 90\%       \\
        Botany                                      & 50                                                   & 50              & 100\%     & 48              & 96\%       \\
        Biotechnology                               & 100                                                  & 74              & 74\%      & 90              & 90\%       \\
        Computer Applications                       & 50                                                   & 40              & 80\%      & 44              & 88\%       \\
        Electrical Engineering                      & 55                                                   & 46              & 84\%      & 49              & 89\%       \\
        Law                                         & 50                                                   & 45              & 90\%      & 42              & 84\%       \\
        Civil                                       & 51                                                   & 46              & 90\%      & 45              & 88\%       \\
        Commerce                                    & 50                                                   & 42              & 84\%      & 49              & 98\%       \\
        Mass Communication                          & 50                                                   & 47              & 94\%      & 40              & 80\%       \\
        Chemistry                                   & 50                                                   & 27              & 54\%      & 37              & 74\%       \\
        Economics                                   & 50                                                   & 50              & 100\%     & 50              & 100\%      \\
        Physiotherapy                               & 64                                                   & 63              & 98\%      & 63              & 98\%       \\
        Optometry                                   & 50                                                   & 45              & 90\%      & 49              & 98\%       \\
        Pharma Sciences                             & 50                                                   & 36              & 72\%      & 32              & 64\%       \\
        Education                                   & 100                                                  & 57              & 57\%      & 75              & 75\%       \\
        Business Management                         & 50                                                   & 41              & 82\%      & 42              & 84\%       \\
        Nutrition and Diet                          & 50                                                   & 42              & 84\%      & 41              & 82\%       \\
        \hline
        Total                                       & 1429                                                 & 1140            & 79\%      & 1245            & 87\%       \\
        \bottomrule
    \end{tabularx}
\end{table*}

%% file: 5_Implications.tex
\section{Implications for Future Research}

The transition from ChatGPT to DeepSeek AI presents new opportunities for researchers to explore advancements in AI, particularly in efficiency, accuracy, and ethical considerations. DeepSeek AI demonstrates improved performance across various domains, leveraging optimized training techniques and better resource management. While ChatGPT has shown strong capabilities in multiple applications, DeepSeek AI consistently achieves higher accuracy, particularly in technical fields such as mathematics. However, generative capabilities and reasoning accuracy remain critical areas of focus, as both models exhibit strengths and limitations in complex problem-solving and creative generation. 

This shift highlights several key areas for future research. One crucial aspect is the development of \textbf{efficient training algorithms} that enable large models to be trained with reduced computational resources, making AI more sustainable and accessible. Additionally, \textbf{multimodal integration} is an important direction, allowing AI systems to process and combine text, audio, and visual inputs for more comprehensive understanding and interaction. Another area of interest is \textbf{continuous learning}, which enables AI models to adapt and improve over time based on user interactions, leading to more personalized and dynamic responses. The accuracy of \textbf{generative AI} also requires further enhancement, ensuring that AI-generated content remains coherent, contextually relevant, and factually accurate. Moreover, \textbf{reasoning capabilities} must be strengthened to allow AI models to provide more reliable and logically sound responses in complex scenarios. 

Furthermore, the need for \textbf{ethical AI development} remains critical, emphasizing the establishment of global standards for fairness, transparency, and bias mitigation. Finally, \textbf{human-AI collaboration} is an emerging field that explores ways to enhance synergy between humans and AI, particularly in creative and decision-making processes. These research directions will shape the future of AI, making systems more efficient, accurate, interactive, and ethically responsible.

%% file: 6_Conclusion.tex
\section{Conclusion}
The evolution from ChatGPT to DeepSeek AI represents a significant milestone in the development of conversational AI. By addressing the limitations of ChatGPT and introducing innovative features, DeepSeek AI sets a new standard for performance, efficiency, and ethical responsibility. Our comparative evaluation also highlights DeepSeek AI's superior performance across multiple domains. As AI continues to evolve, maintaining a focus on transparency, fairness, and responsible development is essential to maximize its social benefits. Future researchers and developers in this field should explore techniques for improving contextual understanding, reducing biases, and optimizing AI efficiency for real-world applications. Furthermore, advances in interpretability and human-AI collaboration will be crucial in making AI systems more reliable and beneficial. The improvements and innovations explored in this paper outline a clear path for future research and progress in artificial intelligence.


\bibliographystyle{unsrt}
\bibliography{reference}